\documentclass[letterpaper, 10 pt, conference]{ieeeconf}  % Comment this line out if you need a4paper

\IEEEoverridecommandlockouts                              % This command is only needed if 
\usepackage{graphicx}
\usepackage{comment}
\usepackage{amsmath,amssymb} % define this before the line numbering.
\usepackage{color}

\usepackage{times}
\usepackage{epsfig}
\usepackage{graphicx}
\usepackage{amsmath}
\usepackage{amssymb}
\usepackage{authblk}

\usepackage[breaklinks=true,bookmarks=false]{hyperref}
\usepackage{paralist,bm} % squeezed itemized lists
\renewcommand{\vec}[1]{\bm{#1}}
\newcommand{\mat}[1]{\bm{#1}}
\newenvironment{packed_itemize}{
\begin{itemize}
  \setlength{\itemsep}{1pt}
  \setlength{\parskip}{0pt}
  \setlength{\parsep}{0pt}
}{\end{itemize}}

\def\eg{\emph{e.g.~}} 
\def\ie{\emph{i.e.~}}

\def\etal{\emph{et al}.~}

\title{\LARGE \bf
$\mathcal{W}$-PoseNet: Dense Correspondence Regularized Pixel Pair Pose Regression
}

\author{Zelin Xu$^{1}$, Ke Chen$^{1,2}$, and Kui Jia$^{1,2}$% <-this % stops a space
\thanks{Z. Xu and K. Chen are equally contributed to this work.}
\thanks{This work was supported by in part by the National Natural Science Foundation of China (Grant No.: 61771201, 61902131), in part by the Program for Guangdong Introducing Innovative and Enterpreneurial Teams
(Grant No.: 2017ZT07X183), the Fundamental Research Funds for the Central Universities (Grant No.: 2019MS022), and the SCUT Program (Grant No.: D6192110).}% <-this % stops a space
\thanks{$^{1}$Z. Xu, K. Chen and K. Jia are with the School of Electronic and
Information Engineering, South China University of Technology, Guangzhou
510641, China {\tt\small \{eezelinxu, chenk, kuijia\}@scut.edu.cn}.}%
\thanks{$^{2}$ K. Chen and K. Jia are also with the Peng Cheng Laboratory, Shenzhen 518066, China. }
}

\begin{document}

\maketitle
\thispagestyle{empty}
\pagestyle{empty}

%%%%%%%%%%%%%%%%%%%%%%%%%%%%%%%%%%%%%%%%%%%%%%%%%%%%%%%%%%%%%%%%%%%%%%%%%%%%%%%%
\begin{abstract}
%Solving 6D pose estimation is non-trivial to cope with intrinsic appearance and shape variations and severe inter-object occlusion, and is made more challenging in light of extrinsic large illumination changes and low quality of the acquired data under an uncontrolled environment. 
This paper introduces a novel 6D pose estimation algorithm -- $\mathcal{W}$-PoseNet, which directly regresses object rotation and translation from a sparse set of pixel pair representations, via a low-rank bilinear pooling on dense features of input RGB-D images. Moreover, those pixel-wise deep features are regularized by explicitly learning a dense correspondence mapping onto their 3D coordinates in a canonical space as an auxiliary task, which can thus improve robustness against ambiguities caused by pose symmetries and inter-object occlusion. Experiment results on two popular benchmarks show that our $\mathcal{W}$-PoseNet consistently achieves state-of-the-art performance on 6D pose estimation. 
\end{abstract}

%%%%%%%%%%%%%%%%%%%%%%%%%%%%%%%%%%%%%%%%%%%%%%%%%%%%%%%%%%%%%%%%%%%%%%%%%%%%%%%%
\section{INTRODUCTION}

The problem of six degree-of-freedom pose (simply put, 6D pose) estimation aims to predicting a rotation together with a translation of an object instance in 3D space relative to a canonical CAD model, which plays a vital role in a number of applications such as augmented reality \cite{marchand2015pose,yu2005pose}, grasp and manipulation in robotics \cite{tremblay2018deep,ten2017grasp,zhu2014single}, and 3D semantic analysis \cite{xu2018pointfusion,tejani2014latent,kehl2016deep,tang2020improving}. 
This paper concerns on the problem of estimating 6D pose of an object given a RGB-D image.
Such a problem remains challenging in view of intrinsic inconsistent texture and shape of objects and inter-object occlusion in the cluttered scenes, in addition to extrinsic varying illumination and sensor noises.
In light of this, encoding a discriminative and robust feature representation for each instance is essential for predicting its 6D pose. 

%Owing to recent success of deep learning in visual recognition \cite{rastegari2016xnor,krizhevsky2012imagenet}, data-driven deep methods are introduced for pose estimation on RGB images via either direct regression on input images to 6D poses \cite{tulsiani2015viewpoints,xiang2015data,sundermeyer2018implicit,mousavian20173d} or exploiting 2D-3D correspondences discovered by detecting 2D projection in RGB images of sparse 3D keypoints \cite{newell2016stacked,oberweger2018making,pavlakos20176,rad2017bb8,peng2019pvnet}.
%Compared to those pose estimation methods relying on object texture extracted from RGB images, RGB-D data can provide extra depth information to mitigate the suffering of lack of textural appearance of an object or fail to extract texture information given low-quality data.

Most of existing RGB-D methods \cite{xiang2017posecnn,sundermeyer2018implicit,kehl2017ssd} depend on iterative-closest point (ICP \cite{besl1992method}) to refine pose predictions, which leads to less efficient inference compared to learning-based refinement (such as the refinement network adopted in \cite{wang2019densefusion} achieving hundreds of magnitude faster) and thus can be less favorable for real-time applications such as robotic grasp and planning.
DenseFusion \cite{wang2019densefusion} has recently been developed to combine both textural and geometric features in a pixel-wise fusion manner, which can obtain accurate pose estimation in a real-time inference processing.
Cheng \etal \cite{cheng20196d} further exploit both intra-modality and inter-modality correlation to learn more discriminative local feature based on DenseFusion \cite{wang2019densefusion}.
However, these state-of-the-art dense pose regressors heavily depend on the quality and resolution of the acquired data and also suffer from pixel-wise feature ambiguities caused by pose symmetries.
%Different from well-established feature learning in the 2D domain \cite{tulsiani2015viewpoints,xiang2015data,sundermeyer2018implicit,mousavian20173d}, learning discriminative feature representations from RGB-D images, containing both texture and geometries of object models, remains open to cope with heavy occlusion, varying texture and shape changes of objects and cross-modality misalignment.

\begin{figure}[t]
\centering
\includegraphics[width=0.8\linewidth]{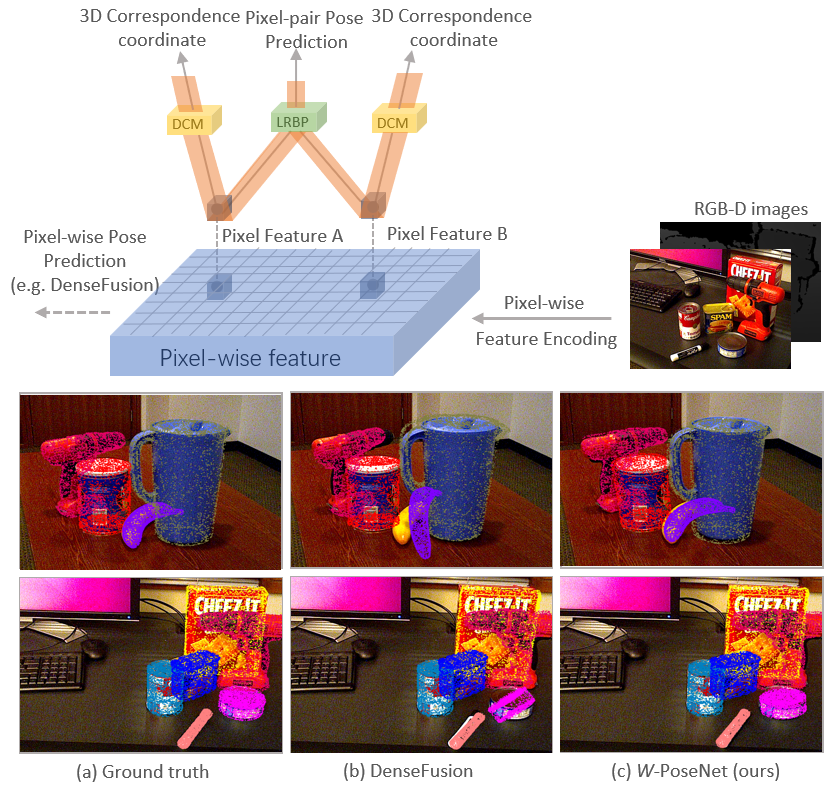}
\caption{Visualization of our $\mathcal{W}$-PoseNet in comparison with its competitor DenseFusion \cite{wang2019densefusion}.
The key differences lie in that our $\mathcal{W}$-PoseNet introduces pixel pair pose estimation based on a low-rank bilinear pooling (LRBP, highlighted in a green block) and dense correspondence mapping (DCM) from each pixel to its 3D coordinate (highlighted in yellow blocks), whose geometric structure is similar to the font $\mathcal{W}$ highlighted in orange. Illustrative examples at the bottom rows are from the YCB-Video \cite{xiang2017posecnn}. }
\label{fig:intro}\vspace{-0.6cm}
\end{figure}

We consider that \textit{extracting discriminative textural and geometric features in local regions} and \textit{generating a robust global representation of each object instance} are important to alleviate the aforementioned challenges in 6D pose estimation.  
To this end, different from pixel-wise regression methods such as DenseFusion \cite{wang2019densefusion} or Correlation Fusion \cite{cheng20196d}, this paper is the first attempt to explore pixel-pair pose regression -- $\mathcal{W}$-PoseNet by developing two novel components of in the proposed deep model: a joint loss function and a $\mathcal{W}$-shape pose regression module, which is shown in Fig. \ref{fig:intro}.

On one hand, beyond one loss term as other pixel-wise pose regressors \cite{wang2019densefusion,cheng20196d} on minimizing pose predictions and their corresponding ground truth, the proposed $\mathcal{W}$-PoseNet additionally encodes the textural and geometric information favoring for reconstructing pose-sensitive point sets, 
whose points' 3D coordinates are generated by transforming the observed point cloud sampling from object models (\eg CAD models) with the inverse of ground truth pose.  
In detail, our $\mathcal{W}$-PoseNet incorporates an auxiliary task of dense correspondence mapping from each pixel of input data to its corresponding coordinate in the canonical space, in order to regularize local feature learning for pose regression by extra pixel-wise supervision self-generated from object models and ground truth pose. 
Intuitively, compared to sparse semantic keypoints pre-defined in keypoint-based methods \cite{newell2016stacked,oberweger2018making,pavlakos20176,rad2017bb8,peng2019pvnet}, dense correspondence mapping in our scheme treats each pixel as a keypoint to regress its corresponding 3D coordinate in object model space, which makes each pixel-wise feature more discriminative, and thus our pixel-wise feature encoding is more robust to occlusion. 

On the other hand, inspired by the point-pair feature \cite{drost2010model,hinterstoisser2016going} on point clouds, this paper proposes a novel pixel pair feature encoding layer based on the low-rank bilinear pooling \cite{kong2017low,wei2018grassmann} to generate pixel-pair features from two sampled pixels' vectors in feature map output of the encoder, which are then mapped onto 6-DoF pose.
Similar to the concept in \cite{drost2010model,hinterstoisser2016going}, our pixel-pair features describe the relative geometric structure of two sampled 3D points on the object models, but also contain texture information of their 2D projection on input RGB images.
Moreover, local features anchored on single pixels concern on extracting textural and geometric information in local regions, which can be less discriminative to some symmetric poses.
Such an observation encourages our motivation to combine pixel-wise features into a set of pixel-pair features.
A global representation for each instance, which combines a number of pixel-pairs' features, thus generates a pose prediction. 
%Soft voting on top confident pose candidates is employed to generate final pose predictions, which can achieve stable and robust performance even given sparsely sampled pixel-pair features.

The main contributions of this paper are three-fold.
\begin{packed_itemize}
\item This paper is the first attempt to introduce the concept of a pair-wise combination of dense features to 6D pose estimation. To this end, this paper proposes a novel pixel-pair pose regression network -- $\mathcal{W}$-PoseNet, which aims to learn a global representation of each object instance consisting of sparse pixel-pair features (PiPF).
\item This paper designs an auxiliary task of dense correspondence mapping from input data to 3D coordinates in object model space sensitive to 6D pose, which can regularize feature encoding in deep pose regression to improve pixel-wise local feature discrimination.
\item Extensive experiments are conducted on the popular benchmarks, whose results can demonstrate that the proposed $\mathcal{W}$-PoseNet consistently outperforms the state-of-the-art 6D pose estimation algorithms. 
\end{packed_itemize}
Source codes and pre-trained models will be released after acceptance at 
%{\url{Link_of_codes_and_pre-trained_models_to_be_released_after_acceptance}}.
{\url{https://github.com/xzlscut/W-PoseNet}}.

%------------------------------------------------------------------------
\section{Related Works}
\begin{figure*}[t]
\centering
\includegraphics[width=0.75\linewidth]{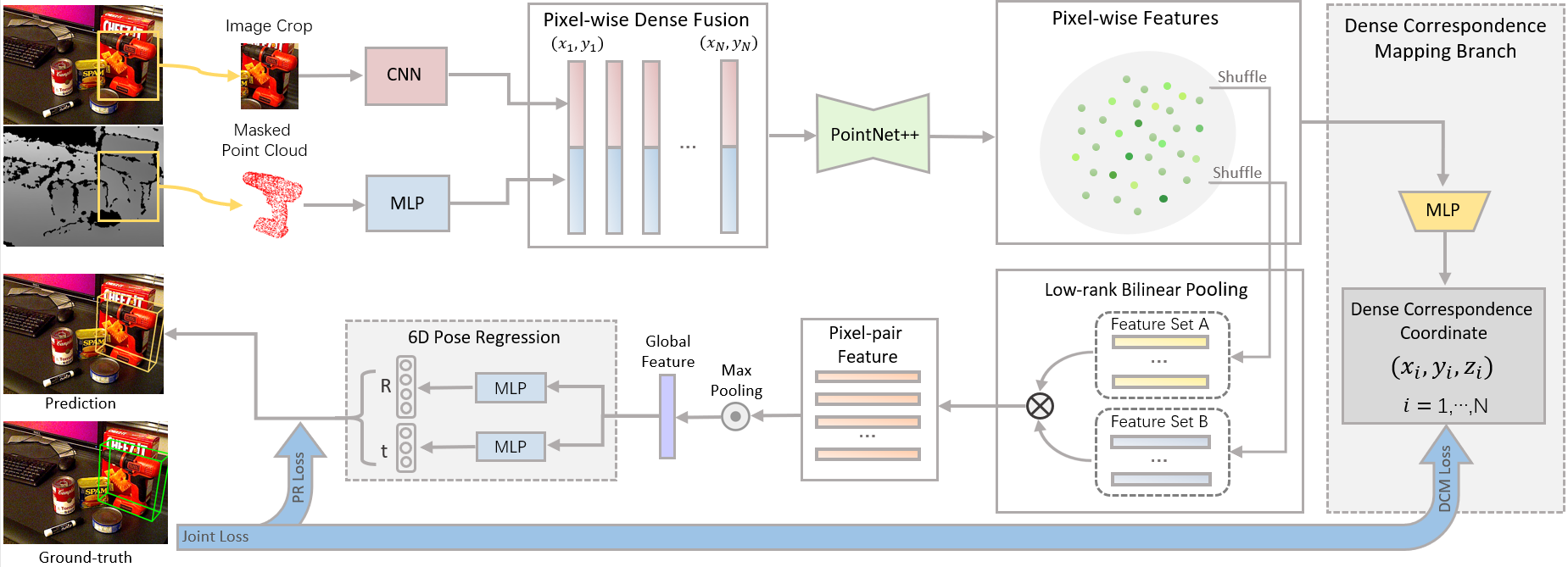}
\caption{Pipeline of the proposed $\mathcal{W}$-PoseNet. The method first detects and segments the foreground containing object instances on RGB images. RGB and depth images are respectively fed into feature encoders and then fused with the PointNet++ \cite{qi2017pointnet++}. Pixel-wise features are sparsely sampled and combined to generate pixel pair features, which produce 6D pose $[\mat{R}|\vec{t}]$. The branch about Dense Correspondence Mapping is to regress pose-specific 3D coordinates from per-pixel features, providing additional geometric constraints for feature learning. A joint loss on dense correspondence mapping and pixel-pair pose regression branches are used to supervise network training. }
\label{fig:pipeline}\vspace{-0.5cm}
\end{figure*}

\vspace{0.1cm}\noindent\textbf{Keypoint-based 6D Pose Estimation --} The algorithms belonging to keypoint-based pose estimation share a two-stage pipeline: first localizing 2D projection of predefined 3D keypoints and then generate pose predictions via 2D-to-3D keypoint correspondence mapping with a PnP \cite{fischler1981random}. 
Existing methods can be categorized into two groups: object detection based \cite{rad2017bb8,tekin2018real} and dense heatmap based \cite{oberweger2018making,pavlakos20176}.
The former focuses on solving the problem of sparse keypoint localization via object detection on the whole object to reduce negative effects of background \cite{rad2017bb8,tekin2018real}, but are sensitive to occlusion \cite{oberweger2018making}. 
The latter group of methods \cite{oberweger2018making,pavlakos20176} pay more attention to discovering latent correlation across all the keypoints and thus are more robust to inter-object occlusion.
Recently, PVNet \cite{peng2019pvnet} is proposed to detect keypoints via voting on pixel-wise predictions of the directional vector that points to keypoints and is robust to truncation and occlusion.
In contrast, dense correspondence mapping in our method share similar concepts as keypoint-based methods but makes pixel-wise predictions on 3D keypoints directly instead of their projection in 2D images, which performs robustly to handle with occlusion. 

\vspace{0.1cm}\noindent\textbf{Dense 6D Pose Estimation --} An alternative group of algorithms is to produce dense pose predictions for each pixel or local patches with hand-crafted features \cite{liebelt2008independent,sun2010depth}, CNN patch-based feature encoding \cite{doumanoglou2016recovering,kehl2016deep} and CNN pixel-based feature encoding \cite{wang2019densefusion,cheng20196d}, whose final pose output is selected via a voting scheme.
In \cite{brachmann2014learning,michel2017global}, random forests are adopted to regress 3d coordinates for each pixel and discover 2D-3D correspondence to produce pose predictions. 
The proposed $\mathcal{W}$-PoseNet is designed to both predict poses from dense features and generate pose-sensitive coordinates in a unique framework.
In general, the dense feature encoding module in our method follows the same pipeline as \cite{wang2019densefusion,cheng20196d}, but the key difference lies in incorporating an extra branch to constrain feature encoding with dense correspondence mapping from input data to its corresponding 3D coordinates in a canonical space.  

\vspace{0.1cm}\noindent\textbf{Bilinear Pooling in CNNs --} The
bilinear pooling \cite{lin2015bilinear} is an effective tool to calculate second-order statistics of local features in visual recognition but suffers from expensive computational cost due to its high-dimensionality.
A number of its variants have been proposed to cope with the challenge via approximation with Random Maclaurin \cite{kar2012random} or Tensor Sketch \cite{pham2013fast}. 
A low-rank bilinear model is proposed by Kong \etal \cite{kong2017low} to avoid computing the bilinear feature matrix, which is approximated via two decomposed low-rank matrices. 
Similarly, \cite{wei2018grassmann} introduces Grassmann Pooling based on SVD decomposition to approximate feature maps.
A quadratic transformation with the low-rank constraint is exploited on pairwise feature interaction either from spatial locations \cite{li2017factorized} or from feature maps across network layers \cite{Yu_2018_ECCV}.  
Our pixel-pair pose regression shares a similar concept as bilinear models to construct pair-wise features, but a sparsely sampled set of pixel-pair features in $\mathcal{W}$-PoseNet can achieve more robust estimation, owing to the pair-wise combination of spatially localized features. 
%Among three combination strategies including element-wise sum, concatenation and low-rank bilinear pooling, the $\mathcal{W}$-PoseNet employing low-rank bilinear pooling achieve the best performance.

%------------------------------------------------------------------------
\section{Methodology}

The problem of 6D pose estimation given RGB-D images is to detect object instances in the scenes and estimate their rotation $\mat{R}\in SO(3)$ and translation $\vec{t}\in  \mathbb{R}^3$. 
In detail, a 6D pose can be defined as a rigid transformation $\vec{p}=[\mat{R}|\vec{t}]$ from the object coordinate system with respect to the camera coordinate system. 
This paper aims to improve the discrimination and robustness of a global representation of each object instance having large variations of texture and shape, from the perspective of pairwise feature interaction on dense local features.

Fig. \ref{fig:pipeline} illustrates the whole pipeline of our $\mathcal{W}$-PoseNet, which consists of several main stages in addition to iterative pose refinement as DenseFusion \cite{wang2019densefusion} (see Sec. \ref{subsec.ipr}).
Encouraged by the recent success of dense pose regression on RGB-D data, we employ the same feature encoding and fusion part of DenseFusion \cite{wang2019densefusion} to extract and fuse pixel-wise features from heterogeneous data (see Sec. \ref{subsec:densefusion}). 
Pixel-wise features are sampled to generate a sparse set of pixel-pair features via a low-rank bilinear pooling on local features, which produces a global feature via a max pooling and its pose prediction (see Sec. \ref{subsec:pppr}).
For robust pixel-wise features, an auxiliary task of learning a dense correspondence mapping from each pixel to its 3D coordinate together with pixel-pair pose regression is designed in Sec. \ref{subsec:dcm}. 
For testing, given an input RGB-D image, the proposed $\mathcal{W}$-PoseNet produces a pose prediction, which as the output of our $\mathcal{W}$-PoseNet is then fed into the iterative pose refinement network to generate a final pose prediction.

%------------------------------------------------------------------------
\subsection{Semantic Segmentation and Pixel-Wise Feature Encoding}\label{subsec:densefusion}

We adopt the identical modules of the DenseFusion \cite{wang2019densefusion} and briefly introduce the following steps: 1) semantic segmentation on RGB images and point clouds converted by depth images; 2) dense feature extraction and 3) pixel-wise feature fusion, which is shown in the top row of Fig. \ref{fig:pipeline}.

Following the segmentation algorithm adopted in \cite{wang2019densefusion,xiang2017posecnn}, an RGB image is first fed into the autoencoder-based segmentation network to produce $N+1$ binary masks belonging to $N$ object classes and the background class respectively.  
The bounding box of each mask is used to crop the corresponding depth image, and the depth intensities within the object mask are converted to a point cloud. 

Cropped image patches and point clouds are fed into 2D CNN based and PointNet \cite{qi2017pointnet} based feature encoders to respectively extract texture and geometric features from heterogeneous data sources.
Specifically, given $H\times W$ RGB-D images, the texture branch aims to mapping $H\times W \times 3$ images to $H\times W \times d_{\text{rgb}}$-dimensional feature maps, which correspond to a feature vector $\vec{f}\in \mathbb{R}^{d_{\text{rgb}}}$ at each spatially localized pixel.
The other branch extracts geometric features $\vec{g}$ from orderless points via two shared-MLP and produces $P \times d_{\text{depth}}$-dimensional feature maps where $P$ denotes the size of sampled points in point cloud generated from the depth image.

%The color feature $\vec{f}$ corresponding to these $P$ points in RGB image patches will then be extracted and used in pixel-wise dense fusion.
We combine textural features $\vec{f}$ from those pixels in cropped RGB image patches corresponding to these $P$ points with $\vec{g}$ in the manner of pixel-wise fusion.
Specifically speaking, to capture correlation across RGB and depth modalities, geometric and appearance features belonging to the same pixels are first concatenated and then fed into a MLP-based feature fusion network (\ie PointNet++ \cite{qi2017pointnet++}) to generate $P$ pixel-wise fusion feature $\vec{x}\in \mathbb{R}^{d_{\text{fusion}}}$.
Instead of concatenating the fused feature with a global feature obtained by average pooling as in the DenseFusion\cite{wang2019densefusion}, we adopted the hierarchical pooling of PointNet++ \cite{qi2017pointnet++} to aggregate local neighboring features.
Our motivation for not using global features here is to get rid of the global information about object instances to verify the effectiveness of a combination of local features as a global description.

%------------------------------------------------------------------------
\subsection{Pixel Pair Pose Regression}\label{subsec:pppr}

This section presents the key components of our pixel-pair pose regression: 1) pixel-pair feature generation; 2) pixel-pair feature encoding; and 3) pose regression, which are shown in the bottom row of Fig. \ref{fig:pipeline}.

%We observe that pixel-wise pose regression in DenseFusion \cite{wang2019densefusion} relies only on appearance and geometric features centered locally on individual pixels; consequently, it may suffer from ambiguities caused by global and/or local pose symmetries.
The seminal work of point-pair feature (PPF) \cite{drost2010model} constructs 4-dimensional features of a relative position in the manner of Euclidean distance and relative angles between normals and between normals and the determined line from a pair of oriented points sampled from an observed 3D scene. 
Motivated from its elegant design that can precisely determine a pose when the pair of points is sampled from the object of interest, we propose to pair-wisely aggregate the learned pixel-wise features for a pixel-pair feature encoding (PiPE) to alleviate the suffering of dense features from ambiguities caused by pose symmetries and occlusion.
%, based on which we learn pixel-pair pose regression. 
%, although the final voting step of weighted average pooling of the confidence scores from individual pixels may alleviate a bit. 
%In contrast, our proposed scheme of PiPF encoding largely avoids the ambiguities in the first place, which in turn further helps the final step of weighted average voting.

\vspace{0.1cm}\noindent\textbf{Generation on Pixel Pair Features --} In 3D domain, local descriptors for point clouds can encode the neighboring geometric structure of each point but suffer from sensor noises and sparse point distribution.
In view of this, multiple point-pair features \cite{drost2010model} consisting of relative position and orientation of two oriented points, are collected as a global representation of an object model.
Such a feature representation can still perform robustly to occlusion and sparse point clouds, which is widely adopted for practical applications.
Inspired by the robustness and low computational cost of point-pair features \cite{drost2010model}, we introduce a novel feature encoding layer inserted between the pose regression module and feature fusion module investigated in the previous section.
Specifically, given the set $\{ \bm{x} \}^P$ of $P$ pixel-wise features, we replicate the features into another set.  
%randomly divide those $\{ \bm{x} \}^P$ feature vectors into two subsets such that points respectively in the two subsets are spatially separated. 
%This practice is to enhance the aforementioned ambiguity alleviation effect of our proposed scheme.
%Specifically, we divided the pixels corresponding to $P$ pixel-wise fusion features into two sets, and let each pixel of one set form a pair with a pixel of another set, and each pixel can only be selected once. 
%After this processing, 
Each feature vector in one set is combined with one randomly selected from the other set, then the $P$ pixels will eventually generate $P$ pixel-pairs.
Other combination strategies such as dense pair generation by combining any two pixels can also be employed, but we adopt such a sparse generation owing to its computational simplicity.

\vspace{0.1cm}\noindent\textbf{Pixel-Pair Feature Encoding (PiPE) --}
Given the output of pixel pairs, a naive approach to combine local features (\textit{i.e.,} $\vec{x}_k$ where $k=1,2,\ldots, P$) into a compact one is direct concatenation, which reveals the 1-order statistical information of local features. 
An alternative to exploit 1-order statistical information is to generate an element-wise sum of two feature vectors.
Encouraged by bilinear models in visual recognition \cite{kar2012random,pham2013fast,wei2018grassmann}, 
our pixel-pair feature encoding is designed based on a low-rank bilinear pooling \cite{kim2016hadamard}.
Specifically, given two pixel-wise feature vectors (\eg $\vec{a}$ and $\vec{b}$ from the set $\{\vec{x}\}^P$), the second-order pooling method \cite{carreira2012semantic} can be written as 
\begin{equation}
\mathcal{G}_{\text{PiPE}}(\vec{a},\vec{b}) = \text{vec}(\vec{a}\vec{b}^T) \in \mathbb{R}^{d_{\text{fusion}}^2}.
\end{equation}
In view of high-dimensionality $d_{\text{fusion}}^2$ of $\vec{a}\vec{b}^T$, the low-rank bilinear pooling \cite{kim2016hadamard} can be adopted to obtain two low-rank matrices $\mat{U}\in \mathbb{R}^{d_{\text{fusion}}\times l}$ and $\mat{V}\in \mathbb{R}^{d_{\text{fusion}}\times l}$ to avoid computing $\vec{a}\vec{b}^T$ as
\begin{equation}
\mathcal{G}_{\text{PiPE}}(\mathbf{a},\mathbf{b}) = \mat{P}^T\sigma(\mat{U}^T\vec{a} \circ \mat{V}^T\vec{b})
\end{equation}
%where $\circ$ denotes the Hadamard product and $\sigma()$ is the activation function which we use relu in our implementation.
where $\circ$ denotes the Hadamard product, $\mat{P}$ is the project matrix to determine the output dimension, and we follow \cite{kim2016hadamard} use the relu nonlinearity activation function $\sigma(\cdot)$.
For generating a $d_{\text{fusion}}$-dimensional pixel-pair feature (the setting in our experiments for a fair comparison), we compare computational complexities of direct concatenation and projection via a fully-connected layer ($2\times d_{\text{fusion}}$, $d_{\text{fusion}}$) and a low-rank bilinear pooling.
The former desires $2\times d_{\text{fusion}}^2$ network parameters, while the latter needs $3\times d_{\text{fusion}}\times l$ where $l$ denoting the dimension of the compressed vector projected from $\vec{x}$ is usually much smaller than $d_{\text{fusion}}$. 
As a result, our method adopts the low-rank bilinear pooling in view of the lighter computational cost and the more richer information than direct concatenation, which is verified in our experiments.

\vspace{0.1cm}\noindent\textbf{Pixel-Pair Pose Regression --}
All $P$ pixel-pair features $\mathcal{G}_{\text{ppf}}$ are fed into a max pooling to generate a global representation for 6D pose prediction of each object instance.
Specifically, the error of pose predictions of the global representation as follows \cite{wang2019densefusion}:
\begin{equation}
  L_{\text{PR}} = \frac{1}{M}\sum_{j}||(\mat{R}x_j + \vec{t}) - (\hat{\mat{R}}x_j + \hat{\mat{t}})||,
\end{equation}
where $x_j$ denotes the $j$-th point in a point set randomly sampled from the object model, $M$ is the size of the point set, $[\mat{R}|\vec{t}]$ is the ground-truth pose and $[\mat{\hat{R}}|\mat{\hat{t}}]$ is the predicted pose output by the pixel-pair feature.
%MLP and each pixel-pair feature generates an object's 6D pose.
%As a result, a set of $P/2$ pose predictions can be collected.
%Similar to the DenseFusion, we used a self-supervision strategy to calculate confidence for each pose prediction. 
%This confidence is then used to generate the final output pose of each object from $P/2$ pose predictions.
%Specifically, self-supervised confidence $c$ are incorporated as an extra term on pose regression loss $L_{\text{PR}}$ which can be depicted as
%\begin{equation}
%  L_{\text{PR}} = \frac{1}{N}\sum_{i}(L^{pair}_{i} c_{i} - w \log(c_{i})),
%\end{equation}
%where  $w$ is a balancing hyper-parameter and $L^{pair}$ denotes the error of pose predictions of each pixel pair feature as follows \cite{wang2019densefusion}:
%\begin{equation}
%L^{pair}_{i} = \frac{1}{M}\sum_{j}\min_{0 < k < M}||(\mat{R}x_j + \vec{t}) - (\hat{\mat{R}_{i}}x_k + \hat{\mat{t}_{i}})||,
%\end{equation}
%where $x_j$ denotes the $j$-th point in a point set randomly sampled from the object model, $M$ is the size of the point set, $[\mat{R}|\vec{t}]$ is the ground-truth pose and $[\mat{\hat{R}}|\mat{\hat{t}}]$ is the predicted pose output by each pixel-pair feature.
%Inspired by the hard voting scheme in 3D point pairs in \cite{drost2010model} for feature matching, we used the confidence-weighted average of $k$ predicted pose to get the pose predictions of our $\mathcal{W}$-PoseNet.

%------------------------------------------------------------------------

\subsection{Auxiliary Dense Correspondence Mapping}\label{subsec:dcm}
\begin{figure}[t]
\centering
\includegraphics[width=0.8\linewidth]{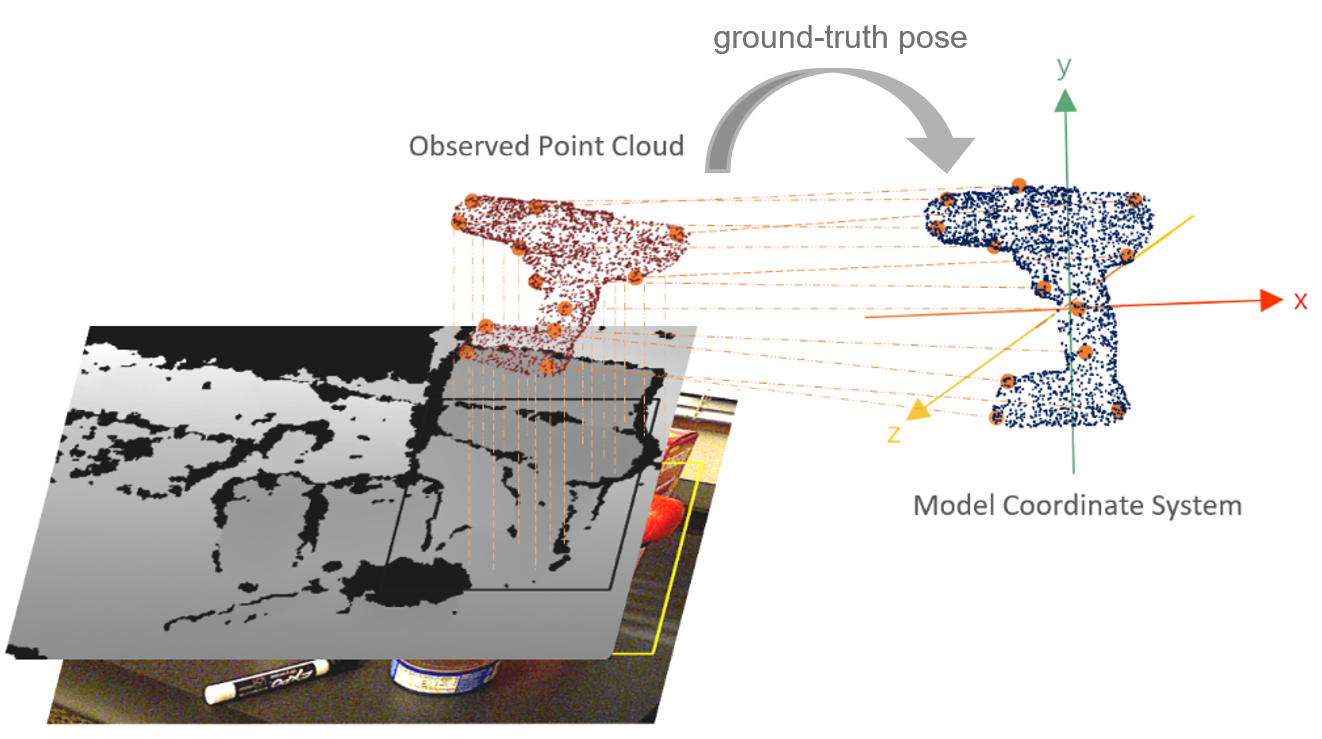}
\caption{Dense Correspondence Mapping (DCM). This module aims to regularizing feature learning in pose regression with mapping onto 3D coordinates sensitive to 6D poses. Specifically, the point cloud in blue is generated by transforming the point clouds sampled from depth image into the model coordinate system, which are used as supervision signals in DCM.}
\label{fig:DCM}\vspace{-0.5cm}
\end{figure}

For obtaining discriminative and robust features per pixels, our method designs an auxiliary dense correspondence mapping task to regularize feature learning.
As shown in Fig. \ref{fig:DCM}, dense correspondence mapping is achieved via regressing per-pixel features $\vec{x}$ to its corresponding 3D coordinate in the canonical coordinate system.
To this goal, the object point cloud obtained by a depth input image is firstly transformed with the inverse of ground truth pose $[\mat{R}|\vec{t}]$ to obtain its corresponding point cloud in the canonical space.
Each 3D coordinate in the transformed point cloud is used as a pixel-wise supervision of its corresponding point in the dense correspondence mapping branch.
The loss term on this auxiliary branch, termed as a Dense Correspondence Mapping (DCM) Loss, can be thus written as  
\begin{equation}
L_{\text{DCM}} = \frac{1}{M}\sum_{i}||p_i - \hat{p}_i||
\end{equation}
where $M$ represents the points sampled from the observed point cloud, $\hat{p}$ is the prediction of the DCM branch of our network, and $p$ denotes the ground truth coordinate corresponding to $\hat{p}$. %which is obtained by converting the coordinates of the observed point cloud in the camera coordinate system to the coordinates in the model coordinate system through gt pose. 
A joint loss $L_{\text{joint}}$ to combine both terms for pose regression and dense correspondence mapping can thus be written as:
\begin{equation}
L_{\text{joint}} = L_{\text{PR}} + \lambda L_{\text{DCM}}
\end{equation}
where $\lambda$ is the trade-off parameter between two terms.

%------------------------------------------------------------------------
\subsection{Iterative Pose Refinement}\label{subsec.ipr}

Iterative refinement in the DenseFusion \cite{wang2019densefusion} has gained significant performance improvement for 6D pose estimation.
%, which is designed based on the concept to correct its pose estimation error in an iterative manner.
Similarly, we adopt its identical refinement post-processing to improve final pose predictions, which encodes the resulting point clouds after transformation with previous pose predictions to gradually refining pose in a residual learning manner. 
Experimental results can demonstrate the boosted performance of the proposed $\mathcal{W}$-PoseNet together with the refinement network.

%The final output pose can be obtained by the concatenation of each iteration's estimation:
%\begin{align}
%\hat{p} = [R_K|t_K] \cdot [R_{K - 1}|t_{K - 1}] \cdot \dots \cdot [R_{0}|t_{0}]
%\end{align}

%------------------------------------------------------------------------
\section{Experiments}
%--------------------------------------------------------------------

\begin{table*}[t]
\begin{center}
\caption{Comparative evaluation of 6D pose estimation on the YCB-Video Dataset in terms of the ADDS AUC and ADD(S) AUC metrics. %We compare the DenseFusion \cite{wang2019densefusion}, the PoseCNN+ICP \cite{xiang2017posecnn} with the proposed $\mathcal{W}$-PoseNet with and without iterative pose refinement in Sec. \ref{subsec.ipr}. 
Objects in bold are symmetric. Results are report in units of $\%$.}
\resizebox{0.9\textwidth}{!}
		{
\begin{tabular}{l|c|c|c|c|c|c||c|c|c|c|c|c} 
\hline
Method                             & \multicolumn{6}{c||}{Without Iterative Refinement}                                                & \multicolumn{6}{c}{With Iterative Refinement}                                                         \\ 
\hline
\multicolumn{1}{c|}{}              & \multicolumn{2}{c|}{PoseCNN \cite{xiang2017posecnn}} & \multicolumn{2}{c|}{DenseFusion \cite{wang2019densefusion}} & \multicolumn{2}{c||}{$\mathcal{W}$-PoseNet} & \multicolumn{2}{c|}{PoseCNN+ICP \cite{xiang2017posecnn}} & \multicolumn{2}{c|}{DenseFusion \cite{wang2019densefusion}} & \multicolumn{2}{c}{$\mathcal{W}$-PoseNet}   \\ 
\hline
\multicolumn{1}{c|}{}              & ADDS & ADD(S)                & ADDS & ADD(S)                    & ADDS & ADD(S)                  & ADDS           & ADD(S)          & ADDS           & ADD(S)          & ADDS\textbf{ } & ADD(S)         \\ 
\hline
002\_master\_chef\_can             & 83.9 & 50.2                  & 95.3 & 70.7                      & 96.0 & 82.2                    & 95.8           & 68.1            & \textbf{96.4} & 73.2            & 96.0           & \textbf{84.0}  \\
003\_cracker\_box                  & 76.9 & 53.1                  & 92.5 & 86.9                      & 93.0 & 87.1                    & 92.7           & 83.4            & \textbf{95.8} & \textbf{94.1}  & 95.5           & 93.0           \\
004\_sugar\_box                    & 84.2 & 68.4                  & 95.1 & 90.8                      & 96.7 & 94.9                    & \textbf{98.2} & \textbf{97.1}  & 97.6           & 96.5            & 97.8           & 96.8           \\
005\_tomato\_soup\_can             & 81.0 & 66.2                  & 93.8 & 84.7                      & 94.3 & 89.0                    & \textbf{94.5} & 81.8            & 94.5           & 85.5            & \textbf{94.5}  & \textbf{89.9}  \\
006\_mustard\_bottle               & 90.4 & 81.0                  & 95.8 & 90.9                      & 97.3 & 95.6                    & \textbf{98.6} & \textbf{98.0}  & 97.3           & 94.7            & 98.1           & 97.5           \\
007\_tuna\_fish\_can               & 88.0 & 70.7                  & 95.7 & 79.6                      & 96.5 & 80.5                    & 97.1           & \textbf{83.9}  & 97.1           & 81.9            & \textbf{97.3}  & 81.8           \\
008\_pudding\_box                  & 79.1 & 62.7                  & 94.3 & 89.3                      & 95.1 & 90.8                    & \textbf{97.9} & \textbf{96.6}  & 96.0           & 93.3            & 96.6           & 94.3           \\
009\_gelatin\_box                  & 87.2 & 75.2                  & 97.2 & 95.8                      & 96.9 & 95.0                    & \textbf{98.8} & \textbf{98.1}  & 98.0           & 96.7            & 98.5           & 97.3           \\
010\_potted\_meat\_can             & 78.5 & 59.5                  & 89.3 & 79.6                      & 90.8 & 78.9                    & \textbf{92.7} & 83.5            & 90.7           & \textbf{83.6}   & 91.6           & 80.4           \\
011\_banana                        & 86.0 & 72.3                  & 90.0 & 76.7                      & 95.8 & 91.2                    & 97.1           & 91.9            & 96.2           & 83.3            & \textbf{97.2}  & \textbf{93.9}  \\
019\_pitcher\_base                 & 77.0 & 53.3                  & 93.6 & 87.1                      & 96.3 & 94.2                    & 97.8           & 96.9            & 97.5           & 96.9            & \textbf{98.3}  & \textbf{98.1}  \\
021\_bleach\_cleanser              & 71.6 & 50.3                  & 94.4 & 87.5                      & 95.2 & 88.6                    & \textbf{96.9} & \textbf{92.5}   & 95.9           & 89.9            & 96.3           & 91.8           \\
\textbf{024\_bowl}                & 69.6 & 69.6                  & 86.0 & 86.0                      & 94.3 & 94.3                    & 81.0           & 81.0            & 89.5           & 89.5            & \textbf{96.2}  & \textbf{96.2}  \\
025\_mug                           & 78.2 & 58.5                  & 95.3 & 83.8                      & 96.5 & 89.7                    & 94.9           & 81.1            & 96.7           & 88.9            & \textbf{97.1}  & \textbf{91.8}  \\
035\_power\_drill                  & 72.7 & 55.3                  & 92.1 & 83.7                      & 95.8 & 93.1                    & \textbf{98.2} & \textbf{97.7}   & 96.0           & 92.7            & 97.4           & 96.2           \\
\textbf{036\_wood\_block}         & 64.3 & 64.3                  & 89.5 & 89.5                      & 91.5 & 91.5                    & 87.6           & 87.6            & \textbf{92.8}  & \textbf{92.8}   & 91.7           & 91.7           \\
037\_scissors                      & 56.9 & 35.8                  & 90.1 & 77.4                      & 88.0 & 60.6                    & 91.7           & 78.4            & \textbf{92.0}  & \textbf{77.9}   & 89.7           & 73.0           \\
040\_large\_marker                 & 71.7 & 58.3                  & 95.1 & 89.1                      & 97.1 & 91.1                    & 97.2           & 85.3            & 97.6           & 93.0            & \textbf{97.5}  & \textbf{90.8}  \\
\textbf{051\_large\_clamp}        & 50.2 & 50.2                  & 71.5 & 71.5                      & 75.7 & 75.7                    & 75.2           & 75.2            & 72.5           & 72.5            & \textbf{76.1}  & \textbf{76.1}  \\
\textbf{052\_extra\_large\_clamp}  & 44.1 & 44.1                  & 70.2 & 70.2                      & 73.3 & 73.3                    & 64.4           & 64.4            & 69.9           & 69.9            & \textbf{74.6}  & \textbf{74.6}  \\
\textbf{061\_foam\_brick}         & 88.0 & 88.0                  & 92.2 & 92.2                      & 95.8 & 95.8                    & \textbf{97.2} & \textbf{97.2}    & 92.0           & 92.0            & 96.9           & 96.9           \\ 
\hline
ALL                                & 75.8 & 59.9                  & 91.2 & 82.9                      & 93.1 & 87.1                    & 93.0           & 85.4            & 93.2           & 86.1            & \textbf{94.1}  & \textbf{89.3}  \\
\hline
\end{tabular}\label{tab.ycb}
}
\end{center}\vspace{-0.5cm}
\end{table*}

%\subsection{Settings}

\vspace{0.1cm}\noindent \textbf{Datasets --}
To evaluate $\mathcal{W}$-PoseNet comprehensively, we conduct experiments on two popular benchmarks -- the YCB-Video dataset \cite{xiang2017posecnn} and the LineMOD dataset \cite{hinterstoisser2011multimodal}.
% {and the OccludedLineMOD dataset \cite{brachmann2014learning}.}
The YCB-Video dataset consists of 21 objects with different textures and sizes and has 92 videos in total. 
Following recent work \cite{xiang2017posecnn,wang2019densefusion,cheng20196d}, we adopt 80 videos for training, 2949 keyframes from the other 12 videos for testing, and also used an additional 80,000 synthetic images provided by \cite{xiang2017posecnn}. 
For a fair comparison, we used the semantic segmentation mask provided by the PoseCNN in evaluation by following \cite{xiang2017posecnn,wang2019densefusion}.
The LineMOD dataset contains 15,783 images belonging to 13 low-textural objects placed under different cluttered environments suffering from the challenges of occlusion and illumination changes. 
We follow the prior works \cite{wang2019densefusion,cheng20196d} use 15\% of images for training and use the other 85\% of images for testing without using additional synthetic data.
%{The OccludedLineMOD dataset provides 6D pose labels of 8 objects in one scene introduced by Hinterstoisser \etal \cite{hinterstoisser2011multimodal}, which includes 1214 images with multiple object instances and are used for testing only, while we use 20,000 synthetic images for training as \cite{peng2019pvnet}.
%Note that, the occluded version of LineMOD benchmark has much more severe inter-object occlusion than its original one (\ie the LineMOD dataset), and is thus made more challenging for evaluation.}

\vspace{0.1cm}\noindent\textbf{Performance Metrics --}
We adopt the ADDS and ADD(S) metrics in the YCB-Video dataset following recent work \cite{xiang2017posecnn,wang2019densefusion} for comparative evaluation. 
For asymmetric objects, we use the average distance (ADD) between 3D model points transformed by ground-truth poses and predicted poses to measure the pose estimation error. 
For symmetric objects, the average closest point distance (ADDS) is employed to measure the mean error. 
6d pose predictions are considered to be correct if the error is smaller than a predefined threshold, which varies from 0 to 10cm to plot an accuracy-threshold curve to compute the area under the curve (AUC).
Similarly, for the LineMOD dataset, we use the ADDS distance for symmetric objects (\ie eggbox and glue) and the ADD distance for the remaining objects having an asymmetric geometry while taking 10\% of the diameter as threshold following \cite{wang2019densefusion,rad2017bb8}.

%, where we adopt the fixed 2cm on theby following recent works \cite{xiang2017posecnn,wang2019densefusion} for a fair comparison . 
% The ADD-S metric treats all objects as symmetric objects adopted in \cite{wang2019densefusion,xiang2017posecnn}, while the ADD(-S) metric employed in \cite{tulsiani2015viewpoints,xiang2015data,sundermeyer2018implicit,mousavian20173d} considers both ADD and ADD-S metrics for asymmetric and symmetric objects respectively.
% Recent works \cite{wang2019densefusion,xiang2017posecnn} employ the fixed 2cm as the threshold in both ADD-S and ADD(-S) on the YCB-Video benchmark, which is the minimum tolerance for robotic manipulation, while adopting 10\% of the diameter for ADD(-S) on the LineMOD and OccludedLineMOD benchmarks.
% For a fair comparison, we use the same setting about pre-defined thresholds in ADD-S and/or ADD(-S) for different datasets.
%Both ADD-S\textless2cm and ADD(-S)\textless2cm metrics consider the pose prediction as correct if the average distance is smaller than 2cm, which is the minimum tolerance for robotic manipulation.

\vspace{0.1cm}\noindent\textbf{Implementation Details --} 
As the DenseFusion \cite{wang2019densefusion}, in our experiments, the CNN used for texture feature encoding is composed of Resnet-18 \cite{he2016deep} followed by 4 blocks of one up-sampling and one convolution layer as the decoder. 
RGB and depth images are respectively encoded into 128-dimensional vectors, which are fused into a pixel-wise dense feature $\vec{x}_{\text{fusion}}$. 
The PointNet++ \cite{qi2017pointnet++} aggregating pixel-wise features consists of two set abstraction and two feature propagation layers. 
For training our model, learning rate is set $10^{-4}$ with the Adam optimizer. 
%23 and 27 epochs are set for training the $\mathcal{W}$-PoseNet and the refinement network respectively, and we refine pose predictions with 2 iterations.
%The refinement network uses 2 iterations in both training and testing.

%------------------------------------------------------------------------
\subsection{Comparison with State-of-the-art Methods}

%--------------------------
\begin{table*}[t]
\begin{center}
\caption{Comparative evaluation of 6D pose estimation in terms of ADD(S) on the LineMOD dataset. Objects in bold are symmetric. The top group of algorithms only use RGB images as input, while the bottom group uses RGB-D images. Ref. denotes the post-processing refinement. Results are reported in units of $\%$.}
\resizebox{.9\textwidth}{!}
		{
\begin{tabular}{l|c|c|c|c|c|c|c|c|c|c|c|c|c|c}
\hline
Method & ape & ben. & cam & can & cat & drill. & duck & \textbf{egg.} & \textbf{glue} & hole. & iron & lamp & phone & \textbf{MEAN} \\ \hline
DeepIM \cite{li2018deepim}& ~77.0 & ~97.5 & ~93.5 & ~96.5 & ~82.1 & ~95.0 & ~77.7 & 97.1 & 99.4 & ~52.8 & ~98.3 & ~97.5 & 87.7 & 88.6 \\
PVNet \cite{peng2019pvnet} & 43.6 & \textbf{99.9} & 86.9 & 95.5 & 79.3 & 96.4 & 52.6 & 99.2 & 95.7 & 82.0 & \textbf{98.9} & 99.3 & 92.4 & 86.3 \\
CDPN \cite{li2019cdpn} & 64.4 & 97.8 & 91.7 & 95.9 & 83.8 & 96.2 & 66.8 & 99.7 & 99.6 & 85.8 & 97.9 & 97.9 & 90.8 & 89.9 \\ \hline
Imp.+ICP \cite{sundermeyer2018implicit} & 20.6 & 64.3 & 63.2 & 76.1 & 72.0 & 41.6 & 32.4 & 98.6 & 96.4 & 49.9 & 63.1 & 91.7 & 71.0 & 64.7 \\
SSD6D+ICP \cite{kehl2017ssd} & 65.0 & 80.0 & 78.0 & 86.0 & 70.0 & 73.0 & 66.0 & \textbf{100.0} & \textbf{100.0} & 49.0 & 78.0 & 73.0 & 79.0 & 79.0 \\
DenseFusion \cite{wang2019densefusion} & 79.5 & 84.2 & 76.5 & 86.6 & 88.8 & 77.7 & 76.3 & 99.9 & 99.4 & 79.0 & 92.1 & 92.3 & 88.0 & 86.2 \\
DenseFusion+Ref. \cite{wang2019densefusion} & 92.3 & 93.2 & 94.4 & 93.1 & 96.5 & 87.0 & 92.3 & 99.8 & \textbf{100.0} & 92.1 & 97.0 & 95.3 & 92.8 & 94.3 \\
$\mathcal{W}$-PoseNet (ours) & 91.7 & 98.8 & 98.4 & 96.5 & 97.7 & 96.3 & 95.0 & 99.8 & 99.9 & 94.4 & 97.7 & \textbf{99.6} & 96.8 & 97.2 \\
$\mathcal{W}$-PoseNet+Ref. (ours) & \textbf{94.9} & 98.9 & \textbf{99.1} & \textbf{97.8} & \textbf{98.8} & \textbf{97.1} & \textbf{97.7} & 99.8 & \textbf{100.0} & \textbf{96.5} & 97.9 & 99.3 & \textbf{97.6} & \textbf{98.1}\\ \hline
\end{tabular}\label{tab.limo}
}
\end{center}\vspace{-0.5cm}
\end{table*}

%--------------------------

We compare the proposed $\mathcal{W}$-PoseNet and the state-of-the-art methods on the YCB-Video and the LineMOD  datasets, and the results are visualized in Tables \ref{tab.ycb} and \ref{tab.limo}. 
In general, our method can consistently achieve state-of-the-art performance with and without iterative pose refinement on the YCB-Video and LineMOD.
Specifically, on the YCB-Video, our network consistently performs better than its direct competitor DenseFusion especially for asymmetric objects with and without post-processing refinement, while similar results are observed on the LineMOD dataset.
As the dense feature extraction and fusion part of our $\mathcal{W}$-PoseNet and DenseFusion are identical, the performance gain can only be explained by our design on a pixel-pair combination of local features and auxiliary dense correspondence mapping. 
% On the more challenging OccludedLineMOD, our $\mathcal{W}$-PoseNet with iterative closest point (ICP) refinement performs slightly better than the PoseCNN+ICP and other comparative methods using RGB-D data, which verifies our model can perform robustly against heavy inter-object occlusion.
%It can be observed from the results on the YCB-Video dataset that our network is more robust against occlusion, especially for 011\_banana and 037\_scissors, which are severely occluded in the test set. On the LineMOD dataset, our method outperforms the state-of-the-art methods in all asymmetric objects.

\begin{figure}[h]
\centering
\includegraphics[width=0.6\linewidth]{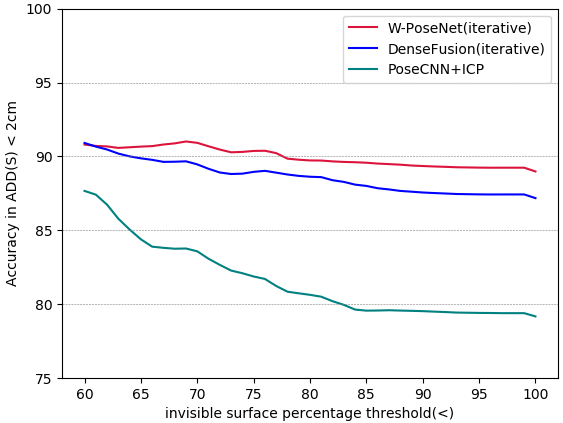}
\caption{Comparison of our $\mathcal{W}$-PoseNet and two state-of-the-art methods under different degree of occlusion.}
\label{fig:occlusion}\vspace{-0.3cm}
\end{figure}

%\vspace{0.1cm}\noindent\textbf{Robustness against Inter-Object Occlusion --} 
To verify the robustness of $\mathcal{W}$-PoseNet against inter-object occlusion, we first calculate the invisible surface percentage for each object instance {on the YCB-Video dataset} as the DenseFusion \cite{wang2019densefusion} did, then evaluate the proposed $\mathcal{W}$-PoseNet with different degrees of occlusion.
Specifically, we sampled a certain number of points on object models, projected these points onto their image plane by using the ground-truth poses and the camera intrinsic parameters. 
Given the observation (\ie depth $d(p)$ on each pixel) in depth images and 2D projection $d(\hat{p})$ of 3D coordinates, if any pixel satisfying $|d(p)-d(\hat{p})|\textgreater{}h$ ($h$ is set to 2cm in our experiment), the pixel is considered under occlusion and can be marked as an invisible pixel. 
The invisible surface percentage for each instance can be generated via the ratio between the size of invisible points and the total number of sampled points.
Fig. \ref{fig:occlusion} shows comparative evaluation of several methods in terms of the ADD(S) metric.
Our $\mathcal{W}$-PoseNet can consistently outperform the PoseCNN+ICP \cite{xiang2017posecnn} and the DenseFusion \cite{wang2019densefusion} with increasing the invisible surface percentage. 
On one hand, both $\mathcal{W}$-PoseNet and DenseFusion are more robust than the PoseCNN+ICP against inter-object occlusion in view of dense feature encoding.
On the other hand, the performance gain of our $\mathcal{W}$-PoseNet sharing the identical feature encoding module of the DenseFusion can demonstrate its superior robustness of our dense correspondence regularized pixel pair pose regression to pixel-wise regression in the DenseFusion.  

%------------------------------------------------------------------------

\subsection{Ablation Studies}
In order to verify the effectiveness of each module of $\mathcal{W}$-PoseNet, we conducted a number of ablation experiments on the LineMOD dataset without iterative pose refinement. 

%Table 2 shows our experimental results. Here, the pixel-pair feature with PR(pose regression) loss refers to replacing the pixel-wise prediction branch of DenseFusion with our pixel-pair prediction branch. The pixel-wise feature with joint loss refers to the addition of a DCM loss on the basis of the DenseFusion network structure.
%\vspace{0.3cm}
\vspace{0.1cm}\noindent\textbf{Effects of Pixel-Pair Pose Prediction --}
Compared to pixel-wise pose estimation in the DenseFusion \cite{wang2019densefusion}, our $\mathcal{W}$-PoseNet without the DCM Loss and post-refining gain 88.2\% on the mean ADD(S), while the DenseFusion without refinement can only achieve 86.2\%.
The only difference between $\mathcal{W}$-PoseNet without the DCM Loss and the DenseFusion lies in the usage of pixel-pair pose regression, and thus 2\% improvement gain can be credited to the effects of Pixel-Pair Pose Prediction.
%The pose prediction based on pixel-pair was 2.0 higher than that based on pixel-wise. This demonstrate that compared to pixel-wise feature, pixel-pair feature provides more global information beneficial to the pose estimation.

\vspace{0.1cm}\noindent\textbf{Evaluation on the DCM Loss --}
We compare the DenseFusion and our $\mathcal{W}$-PoseNet without pixel-pair regression, obtaining 86.2\% and 93.2\% respectively, which only differ on the usage of additional dense correspondence mapping branch.
It is evident that, owing to introducing the auxiliary DCM branch, the $\mathcal{W}$-PoseNet can significantly outperform the state-of-the-art DenseFusion by 7\%, which indicates that the DCM branch improves the quality of per-pixel features.

\begin{table}[h]
\begin{center}
\caption{Ablation studies about combination of local features in terms of ADD(S). 
%ELS, CON, LRBP and DCM respectively denote the element-wise sum, concatenation, low-rank bilinear pooling, and dense correspondence mapping. 
Results are report in units of $\%$.}
\resizebox{0.9\linewidth}{!}
		{
		\begin{tabular}{r c c c || c c c}
\hline
 & ELS & CON  & LRBP & DCM+ELS & DCM+CON  & DCM+LRBP\\ 
\hline
ADD(S) & 91.4 & 92.5 & 92.8 & 96.0 & 97.0 & 97.2\\
\hline		
		\end{tabular}
%\begin{tabular}{llr}
%\hline
%Method &  & ADD(-S) \\ \hline
%DenseFusion \cite{wang2019densefusion} &  & 86.2 \\
%\hline
%$\mathcal{W}$-PoseNet with ELS &  & 91.4 \\
%$\mathcal{W}$-PoseNet with CON &  & 92.5 \\
%$\mathcal{W}$-PoseNet with LRBP &  & 92.8 \\
%\hline
%%DF+PN2+DCM &  & 97.0 \\
%$\mathcal{W}$-PoseNet with ELS+DCM &  & 96.0 \\
%$\mathcal{W}$-PoseNet with CON+DCM &  & 97.0 \\
%$\mathcal{W}$-PoseNet with LRBP+DCM &  & 97.2 \\ \hline
%%DF+PN2+LRBP+DCM+AVR &  & 9 \\ 
%\end{tabular}
}
\end{center}\vspace{-0.5cm}
\end{table}

%\begin{table}[h]
%\begin{center}
%\caption{Ablation studies of different com in terms of ADD(-S) on the LineMOD dataset. Objects with bold name are symmetric. Results are report in units of $\%$.}
%\resizebox{0.35\linewidth}{!}
%		{
%\begin{tabular}{llr}
%\hline
%Method &  & ADD(S) \\ \hline
%DF+PN2 (baseline) &  & 89.9 \\
%DF+PN2+DCM &  & 97.0 \\
%DF+PN2+ELS &  & 91.4 \\
%DF+PN2+CON &  & 92.5 \\
%DF+PN2+LRBP &  & 92.8 \\
%DF+PN2+ELS+DCM &  & 96.0 \\
%DF+PN2+CON+DCM &  & 97.0 \\
%DF+PN2+LRBP+DCM &  & 97.2 \\ \hline
%%DF+PN2+LRBP+DCM+AVR &  & 9 \\ 
%\end{tabular}
%}
%\end{center}\vspace{-0.5cm}
%\end{table}

\vspace{0.1cm}\noindent\textbf{Effects of the Feature Fusion Module --}
{We explore a variety of feature fusion methods, including the element-wise sum (ELS), direct concatenation (CON), and low-rank bilinear pooling (LRBP) under two settings, \ie with and without the auxiliary correspondence mapping (DCM).
Among these methods, the LRBP consistently achieves the best performance in both settings.
Such a study verifies the efficacy of a bilinear pooling to exploit the richer 2-order statistics of local features.}

%BIP and LRBP achieve better performance (96.59\% and 97.02\%) on LineMOD, compared with 95.83\%, 95.59\%, and 93.99\% respectively for CON, ELA, and ELP. Such a study verifies the efficacy of bilinear pooling to exploit the richer information of second-order local features. Among BIP and LRBP, we prefer the lighter and better LRBP.

\vspace{0.1cm}\noindent\textbf{Effects of Iterative Pose Refinement --}
We employ the identical refinement network with the same settings as the DenseFusion.
From Tables \ref{tab.ycb} and \ref{tab.limo}, iterative pose refinement can further boost estimation performance of our $\mathcal{W}$-PoseNet. 
%More importantly, even with iterative refinement, the proposed network can consistently perform better than DenseFusion as well as other competitors.

%\subsection{Visualization}
%Fig. \ref{fig:visualization} shows more qualitative results of the DenseFusion and our $\mathcal{W}$-PoseNet on the YCB-Video dataset. 
%As illustrated in the figure, our method is more robust in cluttered scenes, texture-less objects and occlusion. 
%The fifth column shows an example of a failure due to low-quality mask cropping.
%%the fact that 051\_large\_clamp and 052\_extra\_large\_clamp are completely identical except for their size, making it difficult for the network to distinguish between these two objects.

%------------------------------------------------------------------------

\section{Conclusions}

This paper introduces a novel 6D pose estimation network based on two key observations: \textit{extracting discriminative dense features in local regions} and \textit{generating a robust global representation of each object instance}. 
%To this end, we design two key components based on state-of-the-art DenseFusion, which are verified their effectiveness in the experiments respectively. 
To this end, we introduce two novel modules into the state-of-the-art DenseFusion -- pixel-pair pose prediction and dense corresponding mapping utilizing object models, which are verified their effectiveness in the experiments respectively.
%In addition, we show through comparative experiments that our method is more robust in the face of occlusion.

\bibliographystyle{IEEEtran}
\bibliography{IEEEabrv,6dpose}

\begin{thebibliography}{10}
\providecommand{\url}[1]{#1}
\csname url@rmstyle\endcsname
\providecommand{\newblock}{\relax}
\providecommand{\bibinfo}[2]{#2}
\providecommand\BIBentrySTDinterwordspacing{\spaceskip=0pt\relax}
\providecommand\BIBentryALTinterwordstretchfactor{4}
\providecommand\BIBentryALTinterwordspacing{\spaceskip=\fontdimen2\font plus
\BIBentryALTinterwordstretchfactor\fontdimen3\font minus
  \fontdimen4\font\relax}
\providecommand\BIBforeignlanguage[2]{{%
\expandafter\ifx\csname l@#1\endcsname\relax
\typeout{** WARNING: IEEEtran.bst: No hyphenation pattern has been}%
\typeout{** loaded for the language `#1'. Using the pattern for}%
\typeout{** the default language instead.}%
\else
\language=\csname l@#1\endcsname
\fi
#2}}

\bibitem{marchand2015pose}
E.~Marchand, H.~Uchiyama, and F.~Spindler, ``Pose estimation for augmented
  reality: a hands-on survey,'' \emph{TVCG}, 2015.

\bibitem{yu2005pose}
Y.~K. Yu, K.~H. Wong, and M.~M.-Y. Chang, ``Pose estimation for augmented
  reality applications using genetic algorithm,'' \emph{IEEE Transactions on
  Systems, Man, and Cybernetics, Part B (Cybernetics)}, 2005.

\bibitem{tremblay2018deep}
J.~Tremblay, T.~To, B.~Sundaralingam, Y.~Xiang, D.~Fox, and S.~Birchfield,
  ``Deep object pose estimation for semantic robotic grasping of household
  objects,'' \emph{arXiv preprint arXiv:1809.10790}, 2018.

\bibitem{ten2017grasp}
A.~ten Pas, M.~Gualtieri, K.~Saenko, and R.~Platt, ``Grasp pose detection in
  point clouds,'' \emph{IJRR}, 2017.

\bibitem{zhu2014single}
M.~Zhu, K.~G. Derpanis, Y.~Yang, S.~Brahmbhatt, M.~Zhang, C.~Phillips,
  M.~Lecce, and K.~Daniilidis, ``Single image 3d object detection and pose
  estimation for grasping,'' in \emph{ICRA}, 2014.

\bibitem{xu2018pointfusion}
D.~Xu, D.~Anguelov, and A.~Jain, ``Pointfusion: Deep sensor fusion for 3d
  bounding box estimation,'' in \emph{CVPR}, 2018.

\bibitem{tejani2014latent}
A.~Tejani, D.~Tang, R.~Kouskouridas, and T.-K. Kim, ``Latent-class hough
  forests for 3d object detection and pose estimation,'' in \emph{ECCV}, 2014.

\bibitem{kehl2016deep}
W.~Kehl, F.~Milletari, F.~Tombari, S.~Ilic, and N.~Navab, ``Deep learning of
  local rgb-d patches for 3d object detection and 6d pose estimation,'' in
  \emph{ECCV}, 2016.

\bibitem{tang2020improving}
L.~Tang, K.~Chen, C.~Wu, Y.~Hong, K.~Jia, and Z.~Yang, ``Improving semantic
  analysis on point clouds via auxiliary supervision of local geometric
  priors,'' \emph{arXiv preprint arXiv:2001.04803}, 2020.

\bibitem{xiang2017posecnn}
Y.~Xiang, T.~Schmidt, V.~Narayanan, and D.~Fox, ``Posecnn: A convolutional
  neural network for 6d object pose estimation in cluttered scenes,''
  \emph{arXiv preprint arXiv:1711.00199}, 2017.

\bibitem{sundermeyer2018implicit}
M.~Sundermeyer, Z.-C. Marton, M.~Durner, M.~Brucker, and R.~Triebel, ``Implicit
  3d orientation learning for 6d object detection from rgb images,'' in
  \emph{ECCV}, 2018.

\bibitem{kehl2017ssd}
W.~Kehl, F.~Manhardt, F.~Tombari, S.~Ilic, and N.~Navab, ``Ssd-6d: Making
  rgb-based 3d detection and 6d pose estimation great again,'' in \emph{ICCV},
  2017.

\bibitem{besl1992method}
P.~Besl and N.~D. McKay, ``A method for registration of 3-d shapes,''
  \emph{TPAMI}, 1992.

\bibitem{wang2019densefusion}
C.~Wang, D.~Xu, Y.~Zhu, R.~Mart{\'\i}n-Mart{\'\i}n, C.~Lu, L.~Fei-Fei, and
  S.~Savarese, ``Densefusion: 6d object pose estimation by iterative dense
  fusion,'' in \emph{CVPR}, 2019.

\bibitem{cheng20196d}
Y.~Cheng, H.~Zhu, C.~Acar, W.~Jing, Y.~Wu, L.~Li, C.~Tan, and J.-H. Lim, ``6d
  pose estimation with correlation fusion,'' \emph{arXiv preprint
  arXiv:1909.12936}, 2019.

\bibitem{newell2016stacked}
A.~Newell, K.~Yang, and J.~Deng, ``Stacked hourglass networks for human pose
  estimation,'' in \emph{ECCV}, 2016.

\bibitem{oberweger2018making}
M.~Oberweger, M.~Rad, and V.~Lepetit, ``Making deep heatmaps robust to partial
  occlusions for 3d object pose estimation,'' in \emph{ECCV}, 2018.

\bibitem{pavlakos20176}
G.~Pavlakos, X.~Zhou, A.~Chan, K.~G. Derpanis, and K.~Daniilidis, ``6-dof
  object pose from semantic keypoints,'' in \emph{ICRA}, 2017.

\bibitem{rad2017bb8}
M.~Rad and V.~Lepetit, ``Bb8: A scalable, accurate, robust to partial occlusion
  method for predicting the 3d poses of challenging objects without using
  depth,'' in \emph{ICCV}, 2017.

\bibitem{peng2019pvnet}
S.~Peng, Y.~Liu, Q.~Huang, X.~Zhou, and H.~Bao, ``Pvnet: Pixel-wise voting
  network for 6dof pose estimation,'' in \emph{CVPR}, 2019.

\bibitem{drost2010model}
B.~Drost, M.~Ulrich, N.~Navab, and S.~Ilic, ``Model globally, match locally:
  Efficient and robust 3d object recognition,'' in \emph{CVPR}, 2010.

\bibitem{hinterstoisser2016going}
S.~Hinterstoisser, V.~Lepetit, N.~Rajkumar, and K.~Konolige, ``Going further
  with point pair features,'' in \emph{ECCV}, 2016.

\bibitem{kong2017low}
S.~Kong and C.~Fowlkes, ``Low-rank bilinear pooling for fine-grained
  classification,'' in \emph{CVPR}, 2017.

\bibitem{wei2018grassmann}
X.~Wei, Y.~Zhang, Y.~Gong, J.~Zhang, and N.~Zheng, ``Grassmann pooling as
  compact homogeneous bilinear pooling for fine-grained visual
  classification,'' in \emph{ECCV}, 2018.

\bibitem{qi2017pointnet++}
C.~R. Qi, L.~Yi, H.~Su, and L.~J. Guibas, ``Pointnet++: Deep hierarchical
  feature learning on point sets in a metric space,'' in \emph{NIPS}, 2017.

\bibitem{fischler1981random}
M.~A. Fischler and R.~C. Bolles, ``Random sample consensus: a paradigm for
  model fitting with applications to image analysis and automated
  cartography,'' \emph{Communications of the ACM}, 1981.

\bibitem{tekin2018real}
B.~Tekin, S.~N. Sinha, and P.~Fua, ``Real-time seamless single shot 6d object
  pose prediction,'' in \emph{CVPR}, 2018.

\bibitem{liebelt2008independent}
J.~Liebelt, C.~Schmid, and K.~Schertler, ``independent object class detection
  using 3d feature maps,'' in \emph{CVPR}, 2008.

\bibitem{sun2010depth}
M.~Sun, G.~Bradski, B.-X. Xu, and S.~Savarese, ``Depth-encoded hough voting for
  joint object detection and shape recovery,'' in \emph{ECCV}, 2010.

\bibitem{doumanoglou2016recovering}
A.~Doumanoglou, R.~Kouskouridas, S.~Malassiotis, and T.-K. Kim, ``Recovering 6d
  object pose and predicting next-best-view in the crowd,'' in \emph{CVPR},
  2016.

\bibitem{brachmann2014learning}
E.~Brachmann, A.~Krull, F.~Michel, S.~Gumhold, J.~Shotton, and C.~Rother,
  ``Learning 6d object pose estimation using 3d object coordinates,'' in
  \emph{ECCV}, 2014.

\bibitem{michel2017global}
F.~Michel, A.~Kirillov, E.~Brachmann, A.~Krull, S.~Gumhold, B.~Savchynskyy, and
  C.~Rother, ``Global hypothesis generation for 6d object pose estimation,'' in
  \emph{CVPR}, 2017.

\bibitem{lin2015bilinear}
T.-Y. Lin, A.~RoyChowdhury, and S.~Maji, ``Bilinear cnn models for fine-grained
  visual recognition,'' in \emph{ICCV}, 2015.

\bibitem{kar2012random}
P.~Kar and H.~Karnick, ``Random feature maps for dot product kernels,'' in
  \emph{AISTATS}, 2012.

\bibitem{pham2013fast}
N.~Pham and R.~Pagh, ``Fast and scalable polynomial kernels via explicit
  feature maps,'' in \emph{SIGKDD}, 2013.

\bibitem{li2017factorized}
Y.~Li, N.~Wang, J.~Liu, and X.~Hou, ``Factorized bilinear models for image
  recognition,'' in \emph{ICCV}, 2017.

\bibitem{Yu_2018_ECCV}
C.~Yu, X.~Zhao, Q.~Zheng, P.~Zhang, and X.~You, ``Hierarchical bilinear pooling
  for fine-grained visual recognition,'' in \emph{ECCV}, 2018.

\bibitem{qi2017pointnet}
C.~R. Qi, H.~Su, K.~Mo, and L.~J. Guibas, ``Pointnet: Deep learning on point
  sets for 3d classification and segmentation,'' in \emph{CVPR}, 2017.

\bibitem{kim2016hadamard}
J.-H. Kim, K.-W. On, W.~Lim, J.~Kim, J.-W. Ha, and B.-T. Zhang, ``Hadamard
  product for low-rank bilinear pooling,'' 2016.

\bibitem{carreira2012semantic}
J.~Carreira, R.~Caseiro, J.~Batista, and C.~Sminchisescu, ``Semantic
  segmentation with second-order pooling,'' in \emph{ECCV}, 2012.

\bibitem{hinterstoisser2011multimodal}
S.~Hinterstoisser, S.~Holzer, C.~Cagniart, S.~Ilic, K.~Konolige, N.~Navab, and
  V.~Lepetit, ``Multimodal templates for real-time detection of texture-less
  objects in heavily cluttered scenes,'' in \emph{ICCV}, 2011.

\bibitem{he2016deep}
K.~He, X.~Zhang, S.~Ren, and J.~Sun, ``Deep residual learning for image
  recognition,'' in \emph{CVPR}, 2016.

\bibitem{li2018deepim}
Y.~Li, G.~Wang, X.~Ji, Y.~Xiang, and D.~Fox, ``Deepim: Deep iterative matching
  for 6d pose estimation,'' in \emph{European Conference on Computer Vision
  (ECCV)}, 2018.

\bibitem{li2019cdpn}
Z.~Li, G.~Wang, and X.~Ji, ``Cdpn: Coordinates-based disentangled pose network
  for real-time rgb-based 6-dof object pose estimation,'' in \emph{Proceedings
  of the IEEE International Conference on Computer Vision}, 2019, pp.
  7678--7687.

\end{thebibliography}

\end{document}